\documentclass[10pt,twocolumn,letterpaper]{article}

\usepackage{iccv}
\usepackage{times}
\usepackage{epsfig}
\usepackage{graphicx}
\usepackage{amsmath}
\usepackage{amssymb}
\usepackage{booktabs}
\usepackage{comment}
\usepackage{xcolor}
\usepackage{multirow}
\usepackage{shortbold}
\usepackage{verbatim}
\usepackage{float} %
\usepackage{pifont}
\usepackage{fontawesome5}
\usepackage[accsupp]{axessibility} %
\makeatletter
\@namedef{ver@everyshi.sty}{}
\makeatother
\usepackage{tikz}

\newcommand{\mypar}[1]{\vspace{0.5mm}\noindent\textbf{#1}}

\newcommand{\movableicon}{\faIcon{people-carry}}
\newcommand{\locationicon}{\faIcon{location-arrow}}
\newcommand{\rigidicon}{\faIcon{hammer}}
\newcommand{\artiicon}{\faIcon{door-open}}
\newcommand{\actionicon}{\faIcon{hand-paper}}
\newcommand{\afficon}{\includegraphics[width=0.8em]{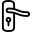}}
\newcommand{\tikzcircle}[2][red,fill=red]{\tikz[baseline=-0.5ex]\draw[#1,radius=#2] (0,0) circle ;}%

\definecolor{msblue}{rgb}{0.310,0.506,0.741}
\definecolor{msgreen}{rgb}{0.608,0.733,0.349}
\definecolor{msred}{rgb}{0.753,0.314,0.302}

\usepackage[pagebackref=true,breaklinks=true,letterpaper=true,colorlinks,bookmarks=false]{hyperref}

\usepackage[capitalize]{cleveref}
\crefname{section}{Sec.}{Secs.}
\Crefname{section}{Section}{Sections}
\Crefname{table}{Table}{Tables}
\crefname{table}{Tab.}{Tabs.}

\iccvfinalcopy %

\def\iccvPaperID{1125} %

\ificcvfinal\fi

\makeatletter
\g@addto@macro\@maketitle{
\vspace{-4.5em}
\begin{center}
    \small \url{https://jasonqsy.github.io/3DOI/}
\end{center}
\vspace{-1.5em}
\begin{figure}[H]
   \setlength{\linewidth}{\textwidth}
\setlength{\hsize}{\textwidth}
\centering
\includegraphics[width=\linewidth]{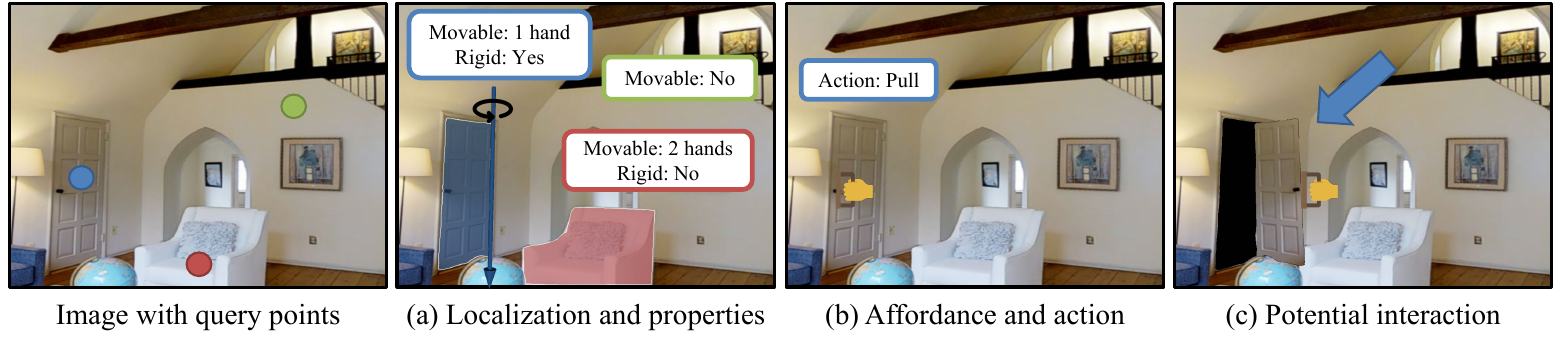}
\caption{Given a single image and a set of query points \tikzcircle[black, fill=msblue]{2.5pt} \tikzcircle[black, fill=msred]{2.5pt} \tikzcircle[black, fill=msgreen]{2.5pt} , our approach predicts: (a) whether the object at the location can be moved \movableicon, its rigidity \rigidicon and articulation class \artiicon, and location \locationicon; (b) an affordance \afficon and action \actionicon; and (c) potential 3D interaction for articulated objects. This ability can assist intelligent agents to better manipulate objects or explore the 3D scene.}
\label{fig:teaser}
\end{figure}
}
\makeatother

\begin{document}

\title{Understanding 3D Object Interaction from a Single Image}

\author{
Shengyi Qian$^\dagger$\\
   $^\dagger$University of Michigan\\
	{\tt\small syqian@umich.edu}\\
\and
David F. Fouhey$^\dagger$$^\ddagger$\\
   $^\ddagger$New York University\\
	{\tt\small david.fouhey@nyu.edu}\\
}

\maketitle
\ificcvfinal\thispagestyle{empty}\fi

\begin{abstract}
Humans can easily understand a single image as depicting multiple potential objects permitting interaction. We use this skill to plan our interactions with the world and accelerate understanding new objects without engaging in interaction.
In this paper, we would like to endow machines with the similar ability, so that intelligent agents can better explore the 3D scene or manipulate objects.
Our approach is a transformer-based model that predicts the 3D location, physical properties and affordance of objects.
To power this model, we collect a dataset with Internet videos, egocentric videos and indoor images to train and validate our approach.
Our model yields strong performance on our data, and generalizes well to robotics data.
\end{abstract}

\section{Introduction}
\label{sec:intro}

What can you do in Figure~\ref{fig:teaser}? This single RGB image conveys a rich, interactive 3D world where you can interact with many objects. For instance, if you grab the chair with two hands, you can move it as a rigid object; the pillow can be picked up freely and squished; and door can be moved, but only rotated. This ability to recognize and interpret potential affordances in scenes helps humans plan our interactions and more quickly learn to interact with objects. The goal of this work is to give the same ability to computers. 

Obtaining such an understanding of potential interactions from a single 3D image is beyond the current state of the art in scene understanding because it spans multiple disparate subfields of computer vision. For instance, single image 3D has made substantial progress~\cite{ranftl2021vision,nie2020total3dunderstanding,yin2021learning,Gkioxari2019}, but primarily focuses on the scene {\it as it exists}, as opposed to {\it as it could be}. There has been an increasing interest in understanding articulation~\cite{li2020category,xiang2020sapien,Qian22}, but these works primarily focus on articulation {\it as it occurs} in a 3D model or carefully collected demonstrations, instead of {\it as it could occur}. Finally, while there is long-standing work on enabling robots to learn interaction and potential interaction points~\cite{pillai2014learning,sturm2011probabilistic}, these works focus primarily on evaluation in primarily the same environment (\eg the lab) and do not focus on applying the understanding in entirely new environments.

We propose to bootstrap this interactive understanding by developing (1) a problem formulation, (2) a rich dataset of annotations on challenging images, and (3) a transformer-based approach. We frame the problem of recognizing the articulation as a prediction-at-a-query-location problem: given an image and  2D location, our method aims to answer ``what can I do here?'' in the style of classic point-and-click games like {\it Myst}. We frame ``what can I do here'' via a set of common questions: whether the object can be moved, its extent when moved and location in 3D, rigidity, whether there are constrains on its motion, as well as estimates of how one would interact the object. To maximize the potential for downstream transfer, our questions are chosen to be generic rather than specific to particular hands or end-effectors: knowing where to act or the degrees of freedom of an object may accelerate reinforcement learning even if one must still learn end-effector-specific skills. 

In order to tackle the task, we introduce a transformer-based model. Our approach, described in Section~\ref{sec:approach} builds on a detection backbone such as Segment-Anything~\cite{kirillov2023segany} in order to build on the advances and expertise of object detection. 
We extend the backbone with additional heads that predict each of our ``what I can I do here'' tasks, and which can be trained end-to-end. As an advantage of our formulation, we can train the system on sparse annotations; we believe this will be helpful for eventually converting our direct supervision to supervision via video.

Powering our approach is a new dataset, described in Section~\ref{sec:dataset}, which we name the 3D Object Interaction dataset ({\it 3DOI}). In order to maximize the likelihood of generalizing to new environments, the underlying data comes from diverse sources, namely Internet and egocentric videos as well as 3D renderings of scene layouts. We provide annotations of our tasks on this data and, due to the source of the data, we also naturally obtain 3D supervision in the form of depth and normals. In total, the dataset has over 50K objects across 10K images, as well as over 31K annotations of non-interactable objects (e.g., floor, wall).

Our experiments in Section~\ref{sec:experiments} test how well our approach recognizes potential interaction, testing on both unseen data in 3DOI as well as robotics data. We compare with a number of alternatives, including generalizing from data of demonstrations~\cite{Qian22,nagarajan2019grounded} and synthetic data~\cite{xiang2020sapien}, as well alternate network designs. Our approach outperforms these models and shows strong generalization to the robotics dataset WHIRL~\cite{bahl2022human}.

To summarize, we see our primary contributions as: (1) the novel task of detecting 3D object interactions from a single RGB image; (2) 3D Object Interaction dataset, which is the first large-scale dataset containing objects that can be interacted and their corresponding locations, affordance and physical properties; (3) A transformer-based model to tackle this problem, which has strong performance on the 3DOI dataset and robotics data.

%
\section{Related Works}

\begin{figure*}[t]
    \centering
    \includegraphics[width=\linewidth]{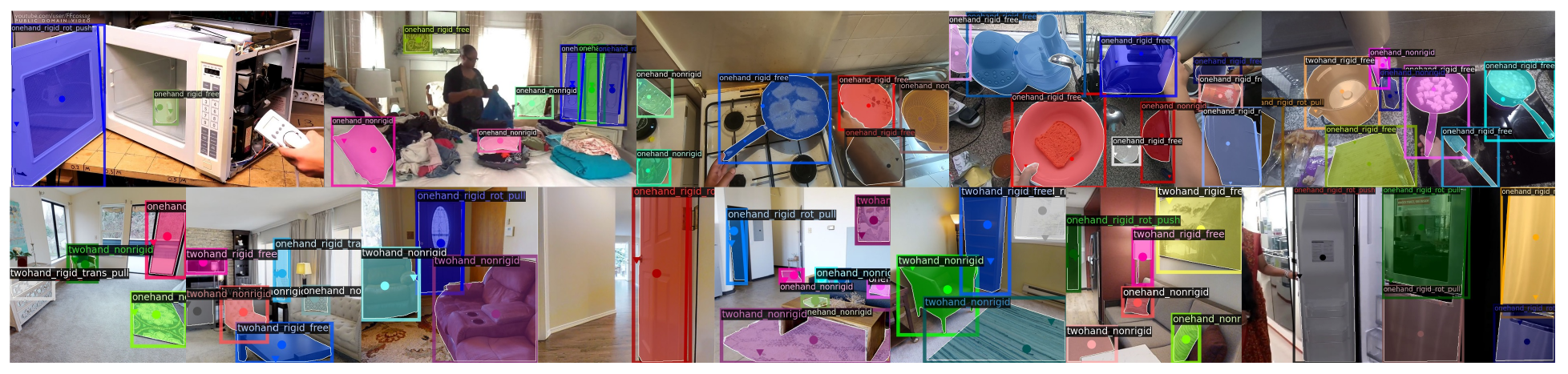}
    \caption{Example annotations of our 3DOI dataset. Our images come from Internet videos~\cite{Qian22}, egocentric videos~\cite{damen2022Rescaling} and renderings of 3D dataset~\cite{eftekhar2021omnidata}. \tikzcircle[black, fill=black]{2pt} is the query point, and $\blacktriangledown$ is the affordance.}
    \label{fig:dataset}
    \vspace{-1em}
\end{figure*}

Our paper proposes to extract 3D object interaction from a single image. This problem lies at the intersection of 3D vision, object detection, human-object interaction and scene understanding.
It is also closely related to downstream robotics applications.

\mypar{Interactive scene understanding.}
Recently, the computer vision community is increasingly interested in understanding 3D dynamics of objects.
It is motivated by human-object interaction~\cite{Chao15,Gkioxari18,Shan20}, although humans do not need to be present in our setting.
Researchers try to understand the 3D shapes, axes, movable parts and affordance on synthetic data~\cite{mo2021where2act,xiang2020sapien,mu2021sdf,jiang2022opd,wei2022self,li2020category,wang2022adaafford}, videos~\cite{Qian22,haresh2022articulated,goyal2022human,nagarajan2019grounded,luo2022learning,li2023locate} or point clouds~\cite{jiang2022ditto,hsu2023ditto}.
Our work is mainly related to \cite{Qian22,haresh2022articulated,goyal2022human} since they work on real images, but is different from them on two aspectives: (1) they need video or multi-view inputs, but our input is only a single image; (2) their approaches recover objects which are being interacted, while our approach understands potential interactions before any interactions happen.
Finally, OPD~\cite{jiang2022opd,sun2023opdmulti} tackles a similar problem for articulated objects, but ours also work for non-articulated objects.

\mypar{Object detection}.
The training anchor-based object detection pipeline basically follows the pipeline of Mask R-CNN~\cite{He17,kirillov2020pointrend,ren2015faster,kirillov2019panoptic}.
As the development of transformer-based models goes, DETR~\cite{carion2020end}, AnchorDETR~\cite{wang2021anchor} and MaskFormer~\cite{cheng2021per} approach object detection as a direct set prediction problem.
Recently, Kirillov \etal proposes Segment Anything Model~\cite{kirillov2023segany}, which predicts object masks from input prompts such as points or boxes.
Our network needs to be built on decoder-based backbones~\cite{carion2020end,cheng2021per,kirillov2023segany}, and we choose SAM~\cite{kirillov2023segany} due to its state-of-the-art performance.

\mypar{Single image 3D.}
Since our problem requires us recover 3D object interaction instead of 2D from a single image, it is also related to single image 3D.
In the recent few years, researchers have developed many different approaches to recover 3D from a single image, including depth~\cite{yin2021learning,ranftl2021vision,Li18,chen2019learning,eigen2015predicting}, surface normals~\cite{Wang15,fan2021three}, 3D planes~\cite{liu2018planenet,liu2019planercnn,jin2021planar} and shapes~\cite{choy20163d,luo2022neural,Gkioxari2019,nie2020total3dunderstanding}.
Our work is built upon their works.
Especially, our architecture is motivated by DPT~\cite{ranftl2021vision} which trains ViT for both segmentation and depth estimation.

\mypar{Robotics manipulation}.
Manipulation of objects is a long-term goal of robotics.
Researchers have developed various solutions for different kinds of objects in different scenes, ranging from articulated objects~\cite{sturm2011probabilistic,pillai2014learning,cifuentes2016probabilistic,desingh2019factored,wu2021vat,gupta2023predicting} to deformable objects~\cite{xu2022dextairity,yang2022touch,wang2021tracking,chi2019occlusion}.
While manipulation is not the goal of our paper, understanding objects and the environment in 3D is typically an important part of a manipulation pipeline.
Our paper mainly improves the perception part, which can potentially improve manipulation.
Therefore, we also test our approach on robotics data~\cite{bahl2022human}, to show it can generalize. %
\section{Overview}
\label{sec:overview}

Given a single image, our goal is to be able to answer ``What could I do here?'' with the object at a query point. We introduce annotations in Section~\ref{sec:dataset} as well as a method for the task in Section~\ref{sec:approach}. Before we do so, we present a unified explanation for the questions we answer as well as the rationale for choosing these questions. We group our questions into six property types, some of which are further subdivided. Not all objects support all questions: objects that cannot be moved, for instance, do not have other properties and objects that can be freely moved do not have rotation axes. We further note that some objects defy these properties -- ball joints, for example, permit a 2D subspace of motion -- our goal is to identify a large subspace of potential interactions.

\mypar{Movable \movableicon}
The most important subdivision is whether the object at the query point can be moved. This follows work in both 3D scene understanding~\cite{silberman2012indoor} and human-object interaction~\cite{Shan20} that subdivide objects into how movable they are. We group objects into three categories based on how easily the object can be moved: (1) {\it fixtures} which effectively cannot be moved, such as walls and floor; (2) {\it one hand} objects that can be moved with a single hand, such as a water bottle or cabinet door; (3) {\it two hand} objects that require two hands to move, such as a large TV. We frame the task as three-way classification.

\mypar{Localization \locationicon}
Understanding the extent of an object is important, and so we localize the object in the world. Since our objects consist of a wide variety of categories, we frame localization as 2D instance segmentation as in~\cite{He17,carion2020end}, as well as a depthmap to localize the object in 3D~\cite{ranftl2021vision,yin2021learning}. These properties can be estimated for most objects.

\mypar{Rigidity \rigidicon} 
To understand action, one primary distinction is rigid-vs-non-rigid since rigid objects are subject to substantially simpler rules of motion~\cite{li2018learning}. We therefore classify whether the object is rigid or not.

\mypar{Articulation \artiicon}
Most rigid objects can further decomposed as permitting freeform, rotational / revolute, or translation / prismatic motion~\cite{sturm2011probabilistic}. Each of these requires different end-effector interactions to effectively interact with. We frame the articulation category as a three-way classification problem, and recognizing the rotation axis as a line prediction problem following~\cite{Qian22}.

\mypar{Action \actionicon}
We also want to understand what the potential action could be to interact with the object. Here we focus on three types of actions: pull, push or other.

\mypar{Affordance \afficon}
Finally, we want to know where we should interact with the object. For example, we need to manipulate the handle if we want to open a door.
We predict a probability map which is over the location of the affordance.

\section{3D Object Interaction Dataset}
\label{sec:dataset}

One critical component of our contribution is accurate annotations of object interactions, as there is no publicly available data.
In this paper, we introduces 3D Object Interaction dataset (3DOI), which is the first dataset.
We picked data that can can be easily integrated with 3D, including a 3D dataset, so that we have accurate 3D ground truth to train our approach.
Examples of our data are shown in Figure~\ref{fig:dataset}.

\mypar{Images.}
Our goal is to pick up diverse images representing real-world scenarios. 
In particular, we want our images contain a lot of everyday objects we can interact with.
Therefore, we sample 10K images from a collection of publicly available datasets:
(1) Articulation~\cite{Qian22} comes from third-person Creative Commons Internet videos. Typically, a video clip contains humans manipulating an articulated objects in households. We randomly sample 3K images from the articulation dataset;
(2) EpicKitchen~\cite{damen2022Rescaling} contains egocentric videos making foods in kitchen environments.
We sample 2K images from EpicKitchen;
(3) Taskonomy~\cite{zamir2018taskonomy} is an indoor 3D dataset with real 2D image and corresponding 3D ground truth. We use the renderings by Omnidata~\cite{eftekhar2021omnidata}. We sample 5k images from the taskonomy split of Omnidata starter dataset.
Overall, there are 10K images.

\begin{figure*}[t]
    \centering
    \includegraphics[width=\textwidth]{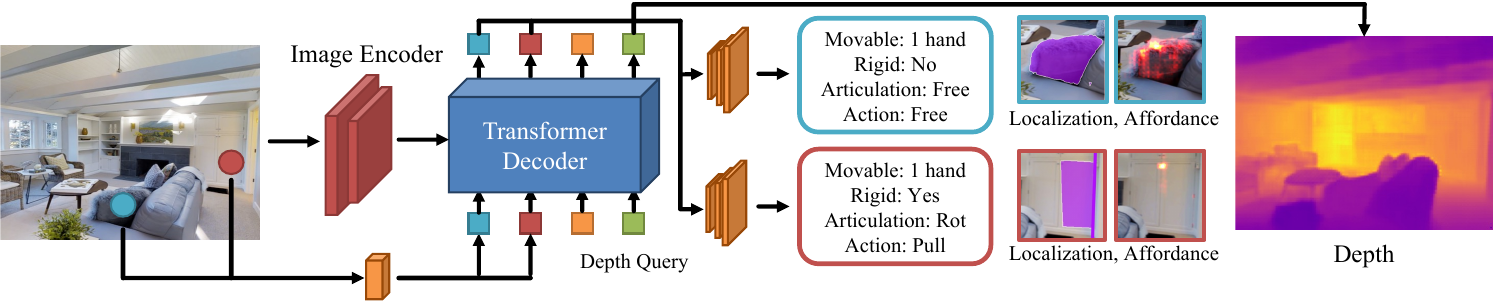}
    \caption{Overview of our approach. The inputs of our network is a single image and a set of query points \tikzcircle[black, fill=msblue]{2.5pt} \tikzcircle[black, fill=msred]{2.5pt}. For each query point, it predicts the potential 3D interaction, in terms of movable \movableicon, location \locationicon, rigidity \rigidicon, articulation \artiicon, action \actionicon and affordance \afficon. In addition, the input of transformer decoder includes a learnable depth query, which estimates the dense depth to recover 3D object interaction for articulated objects.}
    \vspace{-1em}
    \label{fig:approach}
\end{figure*}

\mypar{Annotation.}
With a collection of images with potential objects we can interact, we then turn to manual annotation.
For a single image, we select around $5$ interactable query points, including both large and small objects.
For each query point, we annotate:
(\emph{Movable \movableicon}) one hand, two hand, or fixture.
(\emph{Localization \locationicon}) The bounding box and mask of the part this point belonging to. 
(\emph{Rigidity \rigidicon}) Rigid, or nonrigid.
(\emph{Articulation \artiicon}) Rotation, translation or freeform. We also annotate their rotation axes. 
(\emph{Action \actionicon}) Pull, push or others.
(\emph{Affordance \afficon}) A keypoint which indicates where we should interact with the object.
At the same time, our taskonomy~\cite{zamir2018taskonomy} images come with 3D ground truth, including depth and surface normals.
We also annotate 31K query points of fixtures.
Finally, we split 10K images into a train/val/test set of 8k/1k/1k split, respectively.

\mypar{Availability and Ethics.}
Our images come from three publicly available datasets. 
Taskonomy does not contain any humans. 
The video articulation dataset comes from Creative Commons Internet videos.
We do not foresee any ethical issues in our dataset.
\section{Approach}
\label{sec:approach}

We now introduce a model which can take an image and a set of query point and answer all of questions we asked in Section~\ref{sec:overview}, including movable, localization, rigidity, articulation, action and affordance.
A brief overview of our approach is shown in Figure~\ref{fig:approach}.

Since our inputs include a set of query points and our outputs include both bounding boxes and segmentation masks, we mainly extend SAM~\cite{kirillov2023segany} to build our model.
Compared with traditional detection pipeline such as Mask R-CNN~\cite{He17}, we can use a query point to naturally guide SAM to detect the corresponding object.
Mask R-CNN generates thousands of anchors for each image, which is challenging to find the correct matching.
However, we also compare with alternative network architectures in our experiments for completeness. 
We find they can also work despite being worse than SAM.
For simplicity, we assume there is only a single query point.
But our model can accept hundreds of query points at a time.

\subsection{Backbone}

The goal of our backbone is to map an image $I$ and a query point $[x, y]$ to a pooled feature $h = f(I; [x, y])$.
Full details are in the supplemental.

\mypar{Image Encoder.}
Our image encoder is a MAE~\cite{he2022masked} pretrained Vision Transformer (ViT)~\cite{dosovitskiy2020image}, following SAM~\cite{kirillov2023segany}.
They map a single image $I$ to the memory of the transformer decoder.

\mypar{Query Point Encoder.}
We transfer the query point $[x, y]$ to positional encodings~\cite{tancik2020fourier}, which is then feed into the transformer decoder. 
We use the embedding $k$ to guide the transformer to produce the feature $h$ for different query points.

\mypar{Transformer Decoder.}
The decoder accepts inputs of the memory from the encoder, and an embedding $k$ of the query point.
It produces a embedding $h$ for each query point, and we use it to predict all the properties, like a ROI feature.

\subsection{Prediction Heads}

We now describe how to map from the pooled feature $h$ to the features. Each prediction is done by a separate head that handles each output type.

\mypar{Movable \movableicon} 
We add a linear layer and map the hidden embedding $h$ to the prediction of movable.
We use the standard cross entropy loss to train it.

\mypar{Localization \locationicon}
We follow SAM standard practice to predict segmentation masks.
We predict segmentation masks using mask decoder and train them using focal loss~\cite{lin2017focal} and DICE~\cite{milletari2016vnet} as loss functions.
For depth, we have a separate depth transformer decoder with a corresponding learnable depth query.
We train depth using scale- and shift-invariant L1 loss and gradient-matching loss following~\cite{yin2021learning,ranftl2021vision,Li18}. The shift and scale are normalized per image.

\mypar{Rigidity \rigidicon}
Similar to movable, we add a linear layer to predict whether the object is rigid or not.
We train the linear layer using a standard binary cross entropy loss.

\mypar{Articulation \artiicon}
We first add a linear layer to predict whether the interactive object is rotation, translation or freeform, and we use the standard cross entropy loss to train it.
For the rotation axis, we follow \cite{Qian22,zhao2021deep} to represent an axis as a 2D line $(\theta, r)$. Any points on this line satisfy
$x\cos(\theta) + y\sin(\theta) = r$
where $\theta$ represents the angle and $r$ represents the diatance from the object center to the line.
In training, we represent the 2D line as $(\sin 2\theta, \cos 2\theta, r)$, so that the axis angle is in a continuous space~\cite{zhou2019continuity}.
We use a 3-layer MLP to predict the axis, similar to bounding boxes as both tasks require localization.
We use L1 loss to train it.

\mypar{Action \actionicon}
Similar to movable, we add a linear layer to predict what the potential action is to interact with the object.
We train the linear layer using a standard binary cross entropy loss.

\mypar{Affordance \afficon}
Our prediction of affordance is a probability map, while our annotation is a single keypoint.
However, affordance can have multiple solutions.
Therefore, we transform the annotation of affordance to a 2D gaussian bump~\cite{law2018cornernet} and train the network using a binary focal loss~\cite{lin2017focal}.
We set the weight of positive examples to be 0.95 and that of negative ones to be 0.05 to balance positives and negatives, as there are more negatives than positives.

Our total loss is a weighted linear combination of all losses mentioned above. Details are in supplemental.

\subsection{Implementation Details}
Full architectural details of our approach are in the supplemental. 
In practice, we use three different transformer decoders for mask, depth and affordance.
The image encoder, query point encoder and mask decoder are pretrained on SAM~\cite{kirillov2023segany}.
Other parts, including affordance head and depth head, are trained from scratch.
We use an AdamW optimizer of the learning rate $10^{-4}$, and train our model for 200 epochs.
\section{Experiments}
\label{sec:experiments}

\begin{figure*}[t]
    \centering
    \includegraphics[width=\textwidth]{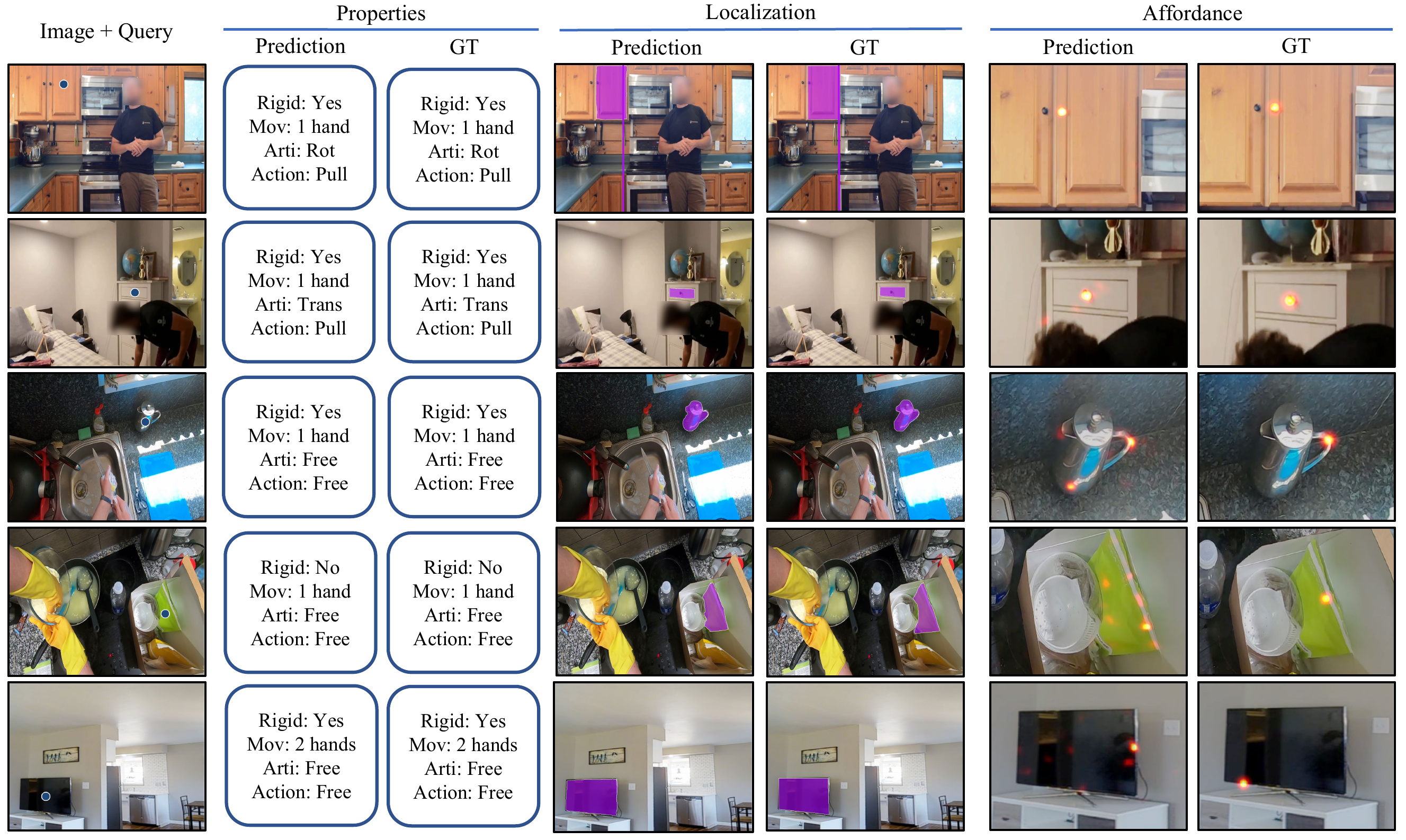}
    \caption{Results on our 3DOI dataset. \tikzcircle[black, fill=msblue]{2.5pt} indicates the query point. \textbf{(Row 1, 2)} Our approach can correctly recognize articulated objects, as well as its type (rotation or translation), axes, and affordance. \textbf{(Row 3, 4)} Our approach can recognize rigid and nonrigid objects in egocentric video. \textbf{(Row 5)} Our approach can recognize objects need to be moved by two hands, such as a TV. We note that the affordance of these objects have multiple solutions. Affordance is zoomed manually for better visualization. Affordance colormap: min \includegraphics[width=4em]{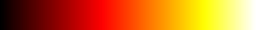}max.}
    \vspace{-0.5em}
    \label{fig:results}
\end{figure*}

We have introduced an approach that can localize and predict the properties of the moving part from an image. 
In the experiments, we aim to answer
the following questions: (1) how well can one localize and predict the properties of the moving part from an image; (2) how well do
alternative approaches to the problem do?
We evaluate our approach on our 3DOI dataset and test the generalization to robotics data WHIRL~\cite{bahl2022human}.

\subsection{Experimental Setup}

We first describe the setup of our experiments. 
Our method aims to look at a single RGB image and infer information about the moving part given a keypoint.
We therefore evaluate our approach on two challenging datasets, using metrics that capture various aspects.

\mypar{Datasets.}
We train and validate our approach on two datasets: 3DOI dataset (described in Section~\ref{sec:dataset}), and the WHIRL dataset~\cite{bahl2022human}. WHIRL~\cite{bahl2022human} is a robotics dataset including every-day objects and settings, for example drawers, dishwashers, fridges in different kitchens, doors to various
cabinets.
We use WHIRL to validate the generalization of our approach and downstream applications in the robotics settings.
We split the first frame of all WHIRL videos and annotate them using the same pipeline as our datasets.
Typically, humans are not present in the first frame and it's before any manipulation.

\mypar{Metrics.}
We report standard practices of evaluation for all of our predictions. For all metrics, the higher the better.
These metrics are detailed as follows:

\noindent
\textbullet\ Movable \movableicon, Rigidity \rigidicon, and Action \actionicon: We report accuracy as these are multiple choice questions.

\noindent
\textbullet\ Localization \locationicon: We report Intersection-over-Union (IoU) for our predictions of bounding boxes and masks~\cite{lin2014microsoft}. We report threshold accuracy for depth~\cite{eigen2015predicting}.

\noindent
\textbullet\ Articulation \artiicon: We report accuracy for articulation type. The rotation axis is a 2D line. Therefore, we report EA-Score between the prediction and the ground truth, following \cite{Qian22,zhao2021deep}. EA-Score~\cite{zhao2021deep} is a score in $[0, 1]$ to measure the angle and euclidean distance between two lines.

\noindent
\textbullet\ Affordance \afficon:
It's a probablity map and we report the histogram intersection (or SIM) following~\cite{nagarajan2019grounded,bylinskii2018different,li2023locate,luo2022learning}.

\mypar{Baselines.}
We compare our approach with a series of baselines, to evaluate how well alternative approaches work on our problem.
We first evaluate 3DADN~\cite{Qian22}, SAPIEN~\cite{xiang2020sapien}, and InteractionHotspots~\cite{nagarajan2019grounded} using their pretrained checkpoints, to test how well learning from videos or synthetic data works on our problem.
We then train two query-point-based model, ResNet MLP~\cite{he2016deep} and COHESIV~\cite{Shan21}, to test how well alternative network architectures work on our problem.
The details are introduced as follows.

\noindent
\textbullet\ (3DADN~\cite{Qian22}): 3DADN detects articulated objects which humans are interacting with, extending Mask R-CNN~\cite{He17}.
It is trained on Internet videos. 
We drop the temporal optimization part since we work on a single image. 
For each image, it can detect articulated objects, as well as the type (translation or rotation), bounding boxes, masks and axes.
Since the inputs of 3DADN do not include a query point, we compare the predicted bounding boxes and the ground truth to find the matching detection, and evaluate other metrics.
We lower the detection threshold to 0.05 to ensure we have enough detections to match our ground truth.

\noindent
\textbullet\ (SAPIEN~\cite{xiang2020sapien}):
The training frames of 3DADN~\cite{Qian22} typically have human activities.
However, our dataset does not require humans to be present, which may lead to generalization issues.
Alternatively, we are interested in whether we can just learn the skill from synthetic data.
We train 3DADN~\cite{Qian22} on renderings of synthetic objects generated by SAPIEN.
SAPIEN is a simulator which contains a large scale set of articulated objects.
We use the renderings provided by 3DADN and the same evaluation strategies.

\noindent
\textbullet\ (InteractionHotspots~\cite{nagarajan2019grounded}): 
While 3DADN and SAPIEN can detect articulated objects as well as their axes, they cannot tell the affordance.
InteractionHotspots learns affordance from watching OPRA~\cite{fang2018demo2vec} or Epic-Kitchen~\cite{damen2022Rescaling} videos.
Since InteractionHotspots cannot detect objects, we apply a center crop of the input image based on the query point, and resize it to the standard input shape (224, 224). 
We use the model trained on Epic-Kitchen as it transfers better than OPRA.

Additionally, we want to test alternative network architectures trained on our 3DOI dataset. 
We use the same loss as ours to train it on 3DOI, to ensure fair comparison.

\noindent
\textbullet\ (ResNet MLP~\cite{he2016deep}): 
ResNet MLP uses a ResNet-50 encoder to extract features from input images.
We then sample the corresponding spatial features from the feature map using the 2D corrdinates of keypoints. 
We train ResNet MLP on all tasks except mask, affordance and depth, as these tasks requires dense predictions for each pixel.
Adding a separate decoder to ResNet makes it a UNet-like architecture~\cite{ronneberger2015unet}, which is beyond the scope of ResNet.

\noindent
\textbullet\ (COHESIV~\cite{Shan21}): 
We also pick another model COHESIV, which designed for the prediction-at-a-query-location problem.
Given an input image and corresponding hand location as a query, COHESIV predicts the segmentation of hands and hand-held objects.
We adopt the network, as it produces a feature map of queries.
we sample an embedding from the feature map according to the query point, concatenate it with image features, and produce multiple outputs.

\begin{figure}[t]
    \centering
     \includegraphics[width=\linewidth]{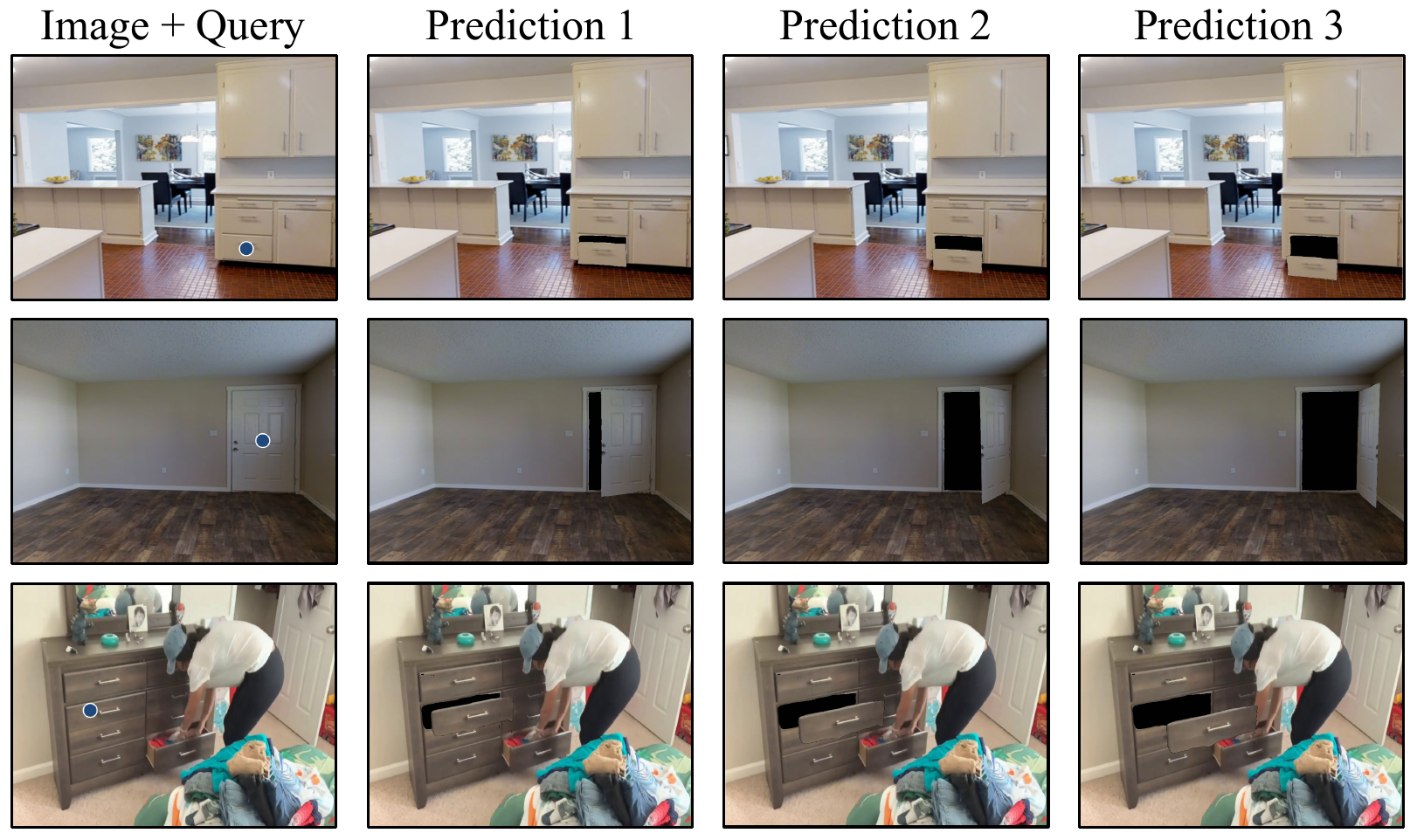}
    \caption{Prediction of 3D potential interaction of articulated objects. \tikzcircle[black, fill=msblue]{2.5pt} indicates the query point. In prediction 1, 2, and 3, we rotate the object along its rotation axis, or translate the object along its normal direction.}
    \label{fig:3darti}
    \vspace{-1em}
\end{figure}

\begin{table*}
   \centering
   \caption{Quantitative results on our 3DOI dataset. Cat. means category. We report accuracy for all category classification, including movable, rigid, articulation and action. We report mean IoU for box and mask, EA-Score for articulation axis, and SIM for affordance. For all metrics, the higher the better.}
   \label{tab:main}
   \scalebox{0.95}{
   \begin{tabular}{lccccccccc}
      \toprule
       & Movable \movableicon & \multicolumn{2}{c}{Localization \locationicon} & Rigidity \rigidicon & \multicolumn{2}{c}{Articulation \artiicon} & Action \actionicon & Affordance \afficon \\
       \cline{3-4} \cline{6-7}
      Methods & Cat. & Box & Mask & Cat. & Cat. & Axis & Cat. & Probability\\
      \midrule
      3DADN~\cite{Qian22} & - & 8.53 & 6.45 & - & 44.3 & 5.63 & - & - \\
      SAPIEN~\cite{xiang2020sapien} & - & 5.94 & 4.57 & - & 41.6 & 1.79 & - & - \\
    InteractionHotspots~\cite{nagarajan2019grounded} & - & - & - & - & - & - & - & 0.047 \\
    ResNet MLP~\cite{he2016deep} & 72.5 & 21.4 & - & 81.9 & 51.9 & 68.3 & 58.8 & - \\
      COHESIV~\cite{Shan21} & 71.5 & 28.3 & 35.2 & 81.2 & 68.0 & 67.2 & 71.5 & 0.013 \\
      \textbf{Ours} & \textbf{85.3} & \textbf{69.9} & \textbf{77.1} & \textbf{90.1} & \textbf{89.4} & \textbf{80.3}  & \textbf{89.7} & \textbf{0.167} \\
      \bottomrule
   \end{tabular}
   } %
   \vspace{-1em}
\end{table*}

\subsection{Results}

First, we show qualitative results in Figure~\ref{fig:results}. 
For articulated objects (drawers, cabinets, etc.), our approach can recognize its location, kinematic model (rotation or translation), axes and handle. It can also recognize rigid or nonrigid objects, as well as light or heavy ones.
It works on both third-person images or egocentric videos.
And all of these are achieved in a single model.
For articulated objects, we utilize the outputs and further show their potential 3D interaction in Figure~\ref{fig:3darti}. Full details in supplemental.

\begin{figure}[t]
    \centering
    \includegraphics[width=\linewidth]{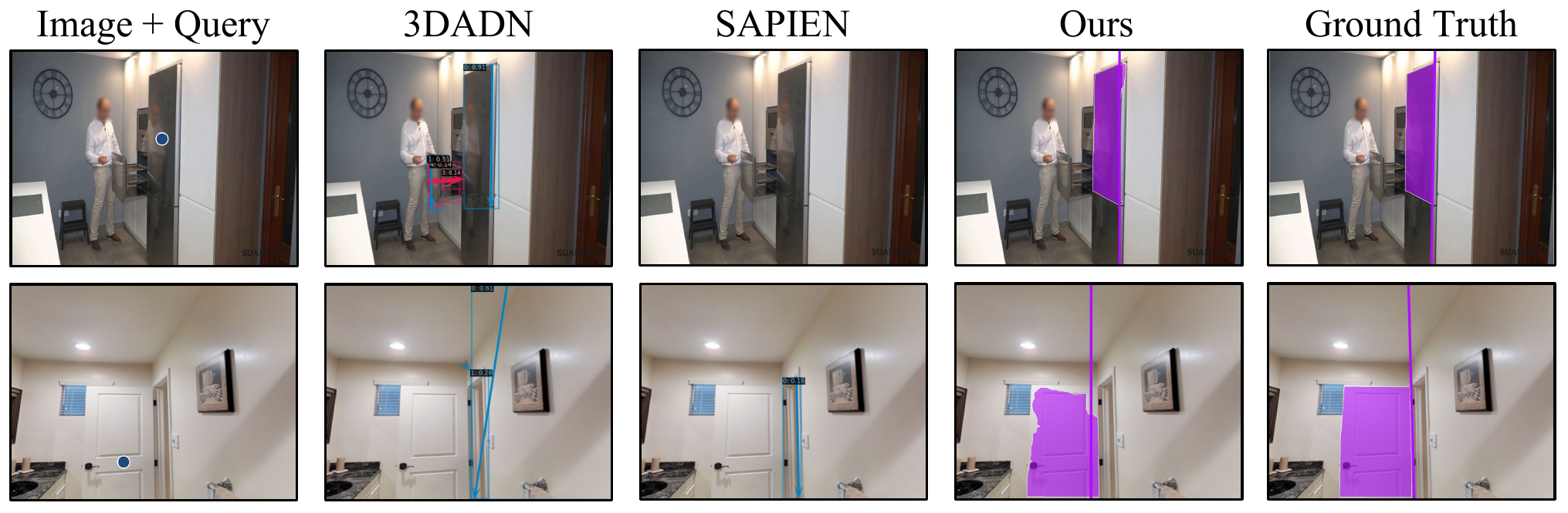}
    \caption{Comparison of 3DADN~\cite{Qian22}, SAPIEN~\cite{xiang2020sapien} and our approach. \tikzcircle[black, fill=msblue]{2.5pt} indicates the query point. 3DADN has a strong performance when humans are present. However, it has difficulty detecting objects without human activities. SAPIEN does not generalize well to real images. However, it is sometimes better than 3DADN when humans are not present.}
    \vspace{-1em}
    \label{fig:3dadn}
\end{figure}

We then compare our approach with a set of baselines.
The quantitative results are reported in Table~\ref{tab:main}.
3DADN~\cite{Qian22} is much worse than our approach, since it can only detect objects which are being articulated.
It fails to detect objects humans are not interacting. 
Instead, our approach can detect any objects can be interacted, regardless of human activities.
SAPIEN is worse than 3DADN, which suggests learning from synthetic objects has a huge domain gap.
This is consistent with the observation of 3DADN.
Visual comparisons are shown in Figure~\ref{fig:3dadn}.

\begin{figure}[t]
    \centering
    \includegraphics[width=\linewidth]{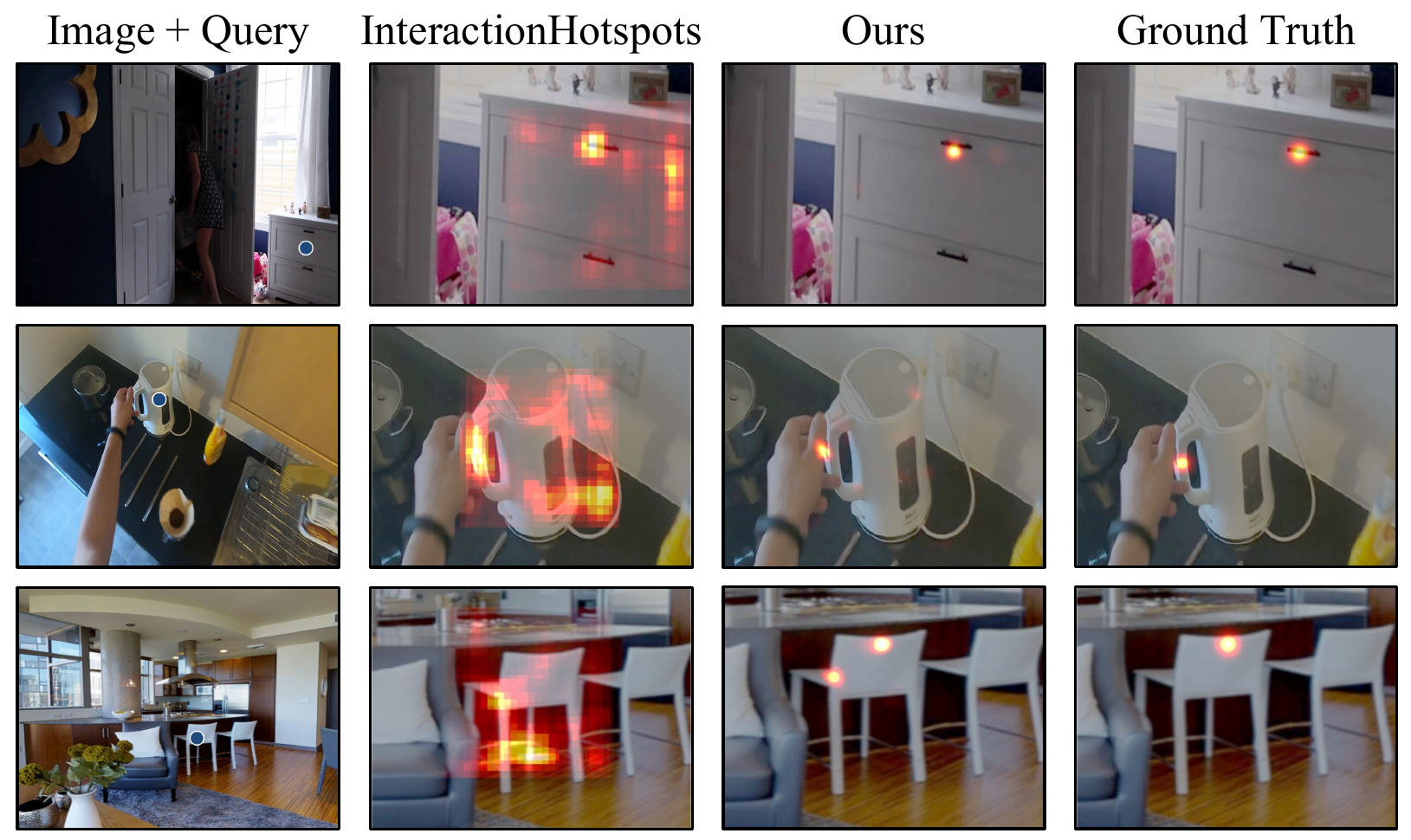}
    \caption{Comparison of InteractionHotspots~\cite{nagarajan2019grounded} and our approach. \tikzcircle[black, fill=msblue]{2.5pt} indicates the query point. We find InteractionHotspots typically makes a cloud like probability map on our data. Our model is very confident about its prediction, while there can be multiple solutions. Prediction and GT are zoomed manually for better visualization. Affordance colormap: min \includegraphics[width=5em]{figures/colorscale_hot.jpg} max.}
    \vspace{-1em}
    \label{fig:hotspots}
\end{figure}

We compare our prediction of the affordance map with InteractionHotspots~\cite{nagarajan2019grounded}.
Our approach outperforms InteractionHotspots significantly, with a 3.5x improvement.
A visual comparison is shown in Figure~\ref{fig:hotspots}.
While InteractionHotspots predicts a cloud-like probability map, our approach is typically very confident about its prediction.
However, the overall performance is relatively low, mainly due to ambiguity of affordance on deformable objects.

To explore alternative network architectures, we compare our approach with ResNet MLP~\cite{he2016deep} and COHESIV~\cite{Shan21}, which are trained on our data with the same loss functions.
ResNet MLP is reasonable on movable, rigidity, and action.
It is especially bad on bounding box localization, which is why we typically rely on a detection pipeline such as Mask R-CNN~\cite{He17}.
COHESIV learns reasonable bounding boxes and masks, which is a huge improvement over ResNet MLP.
The performance of movable drops compared with ResNet MLP, while that of kinematic and action improves.
Overall, our approach outperforms both ResNet MLP and COHESIV, mainly due to the introduction of transformers.

Finally, we evaluate depth on our data.
Having state-of-the-art depth estimation is orthogonal to our goal, since we only need reasonable depth to localize objects in 3D and render potential 3D interactions.
In fact, state-of-the-art depth estimation models are trained on over ten datasets and one million images~\cite{ranftl2021vision,yin2021learning,eftekhar2021omnidata}, while our dataset only has 5K images with depth ground truth.
We just report the evaluation of depth estimation, in order to show our model has learned reasonable depth.
On our data, 96.7\% pixels are within the $1.25$ threshold, 99.3\% pixels are within the $1.25^2$ threshold.

\begin{table*}
   \centering
   \caption{Quantitative results on robotics data~\cite{bahl2022human}. Cat. means category. We report accuracy for all category classification, including movable, rigid, articulation and action. We report mean IoU for the boxes and masks, EA-Score for articulation axis, and SIM for affordance probablity map. For all metrics, the higher the better.}
   \label{tab:whirl}
   \scalebox{0.95}{
   \begin{tabular}{lccccccccc}
      \toprule
       & Movable \movableicon & \multicolumn{2}{c}{Localization \locationicon} & Rigidity \rigidicon & \multicolumn{2}{c}{Articulation \artiicon} & Action \actionicon & Affordance \afficon \\
       \cline{3-4} \cline{6-7}
      Methods & Cat. & Box & Mask & Cat. & Cat. & Axis & Cat. & Probability\\
      \midrule
      3DADN~\cite{Qian22} & - & 13.8 & 10.1 & - & 53.3 & 4.03 & - & - \\
      SAPIEN~\cite{xiang2020sapien} & - & 9.14 & 6.15 & - & 51.1 & 0.0 & - & - \\
    InteractionHotspots~\cite{nagarajan2019grounded} & - & - & - & - & - & - & - & 0.045 \\
    ResNet MLP~\cite{he2016deep} & 88.8 & 14.1 & - & 80.0 & 51.1 & 57.1 & 51.1 & - \\
      COHESIV~\cite{Shan21} & 86.7 & 37.1 & 38.7 & 82.2 & 73.3 & 66.1 & 73.3 & 0.015 \\
      \textbf{Ours} & \textbf{91.1} & \textbf{68.7} & \textbf{70.2} & \textbf{95.6} & \textbf{80.0} & \textbf{68.5}  & \textbf{84.4} & \textbf{0.148} \\
      \bottomrule
   \end{tabular}
   } %
   \vspace{-1em}
\end{table*}

\begin{figure}[t]
    \centering
    \includegraphics[width=\linewidth]{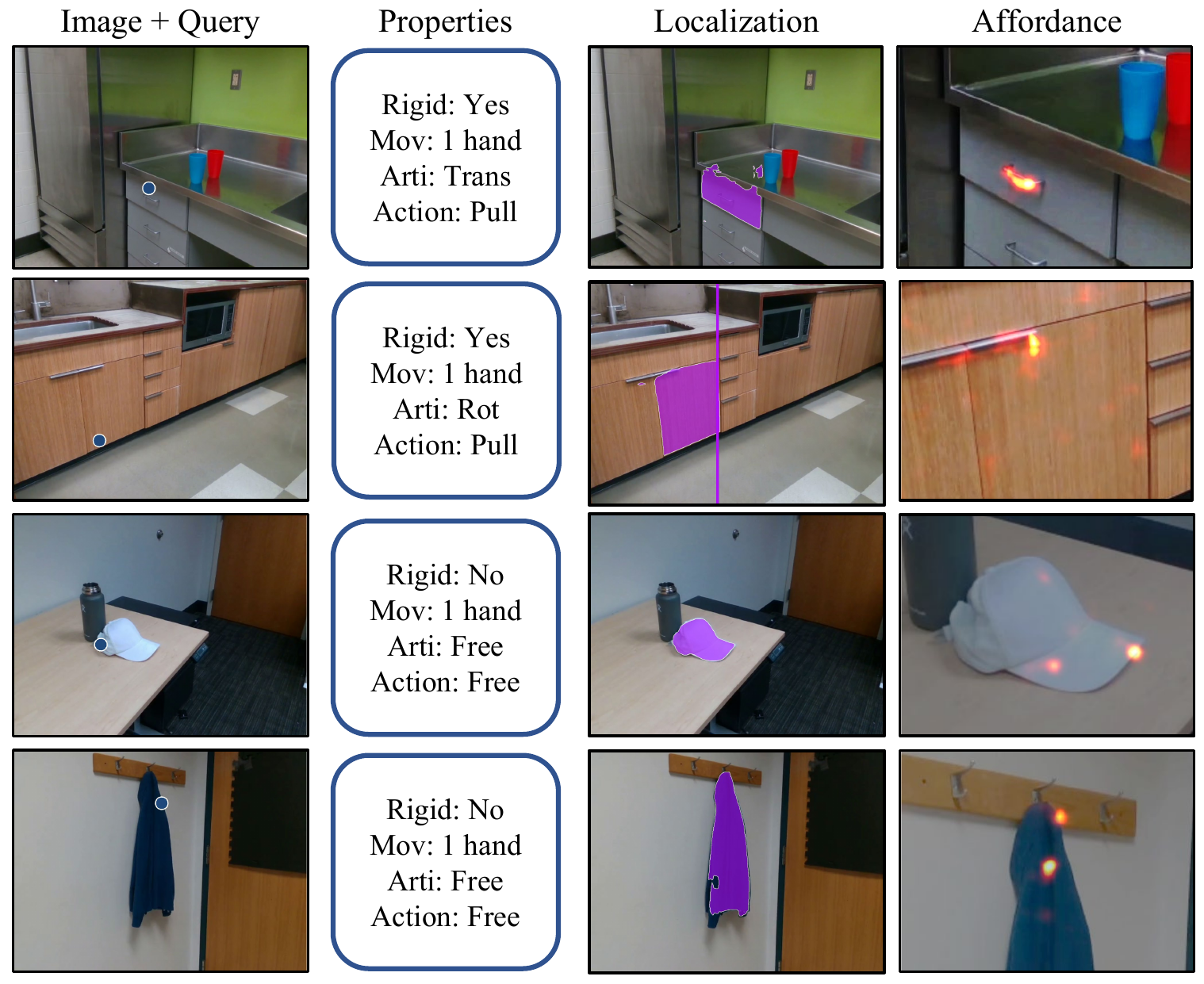}
    \caption{Results on robotics data~\cite{bahl2022human}. \tikzcircle[black, fill=msblue]{2.5pt} indicates the query point. Without finetuning, our approach generalizes well to robotics data, which indicates its potential to help intelligent agents to better manipulate objects. Row 1 and 2 are articulated objects. Row 3 and Row 4 are deformable objects. Affordance is zoomed manually for better visualization. Affordance colormap: min \includegraphics[width=5em]{figures/colorscale_hot.jpg}max.}
    \vspace{-1.5em}
    \label{fig:whirl}
\end{figure}

\subsection{Generalization Results}

To test whether our approach and models trained on our 3DOI dataset can generalize,
we further evaluate our approach on WHIRL~\cite{bahl2022human}, a robotics dataset manipulating every-day objects.
Since WHIRL is a small-scale dataset, we test our model on WHIRL without finetuning.
Our results are shown in Figure~\ref{fig:whirl}. 
For both articulated objects and deformable objects, our approach can successfully recover its kinematic model, location and affordance.

We also quantitatively evaluate our approach on WHIRL. We report our results in 
Table~\ref{tab:whirl}.
Similar to our 3DOI dataset, our approach outperforms 3DADN~\cite{Qian22}, SAPIEN~\cite{xiang2020sapien} and InteractionHotspots~\cite{nagarajan2019grounded} significantly.
The performance gap is even larger.
We believe it is because humans are not present in most images of the dataset.

We compare our approach with ResNet MLP~\cite{he2016deep} and COHESIV~\cite{Shan20}, which are also trained on our 3DOI dataset.
Our model outperforms both ResNet MLP and COHESIV consistently.
The improvement on dense predictions (Localization and Affordance) is significant, due to the design of mask decoder. 
The improvement on other properties is relatively small.
It illustrates models trained on our 3DOI dataset generalize well to robotics data, regardless of network architectures.

\begin{figure}[t]
    \centering
    \includegraphics[width=\linewidth]{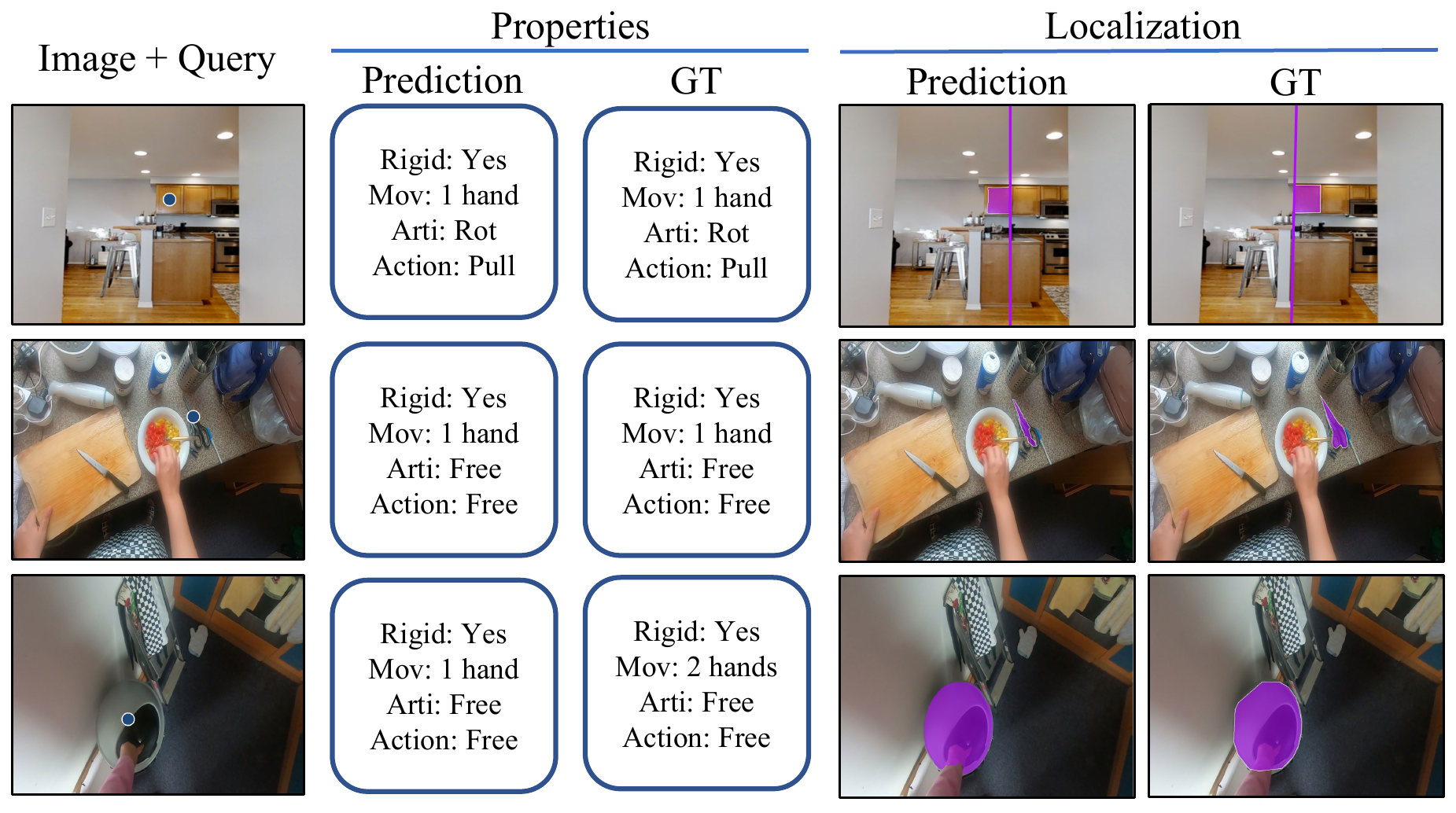}
    \caption{Typical failure modes of our approach. \tikzcircle[black, fill=msblue]{2.5pt} indicates the query point. \textbf{Row 1: } Our predicted rotation axis is on the wrong side when the objects look symmetric. \textbf{Row 2: } Our predicted mask is partial when the scissors are occluded. \textbf{Row 3: } Our model thinks the trash bin can be picked up by 1 hand, potentially since its material looks plastic.}
    \vspace{-1.5em}
    \label{fig:failure}
\end{figure}

\subsection{Limitations and Failure Modes}

We finally discuss our limitations and failure modes.
In Figure~\ref{fig:failure}, we show some predictions are hard to make from visual cues:
Some articulated objects are symmetric and humans rely on common sense to guess its rotation axis.
There are also hard examples when predicting the rigidity and movable. 
Finally, we only annotate a single keypoint for each object instance as affordance.
But some objects may have multiple keypoints as affordance.
 
\section{Conclusion}
We have presented a novel task of predicting 3D object interactions from a single RGB image.
To solve the task, we collected the 3D Object Interaction dataset, and proposed a transformer-based model which predicts the potential interactions of any objects according to query points.
Our experiments show that our approach outperforms existing approaches on our data and generalizes well to robotics data. 

Our approach can have positive impacts by helping build smart robots that are able to understand the 3D scene and manipulate everyday objects. On the other hand, our approach may be useful for surveillance activities.

\mypar{Acknowledgments}
This work was supported by the
DARPA Machine Common Sense Program.
This material is based upon work supported by the National Science Foundation under Grant No. 2142529.
We thank Shikhar Bahl and Deepak Pathak for their help with WHIRL data, Georgia Gkioxari for her help with the figure, and Tiange Luo, Ang Cao, Cheng Chi, Yixuan Wang, Mohamed El Banani, Linyi Jin, Nilesh Kulkarni, Chris Rockwell, Dandan Shan, Siyi Chen for helpful discussions.

{\small
\bibliographystyle{ieee_fullname}
\bibliography{local}
}

\appendix

\clearpage
\section{Implementation}
\label{app:implementaion}

\mypar{Transformer Decoder.}
The transformer decoder $D$ takes the memory $m$ from encoder and a set of queries, including $N$ point queries $k_p$ and one depth query $k_d$.
It predicts a set of point pooled features $h_1, \dots, h_{N}$ and depth pooled features $h_d$, \ie
\begin{equation}
    h_1, h_2,\dots, h_{N}, h_d = D(m; k_p^{(1)}, k_p^{(2)},\dots k_p^{(N)}, k_d)
\end{equation}
We set $N = 15$, as all images have lower than 15 query points.
For images without 15 query points, we pad the input to 15 and do not train on these padding examples.
The depth query $k_d$ is a learnable embedding, similar to object queries in DETR~\cite{carion2020end}.
All queries are feed into the decoder in parallel, as they are indepedent of each other.

\mypar{Prediction heads.}
DETR~\cite{carion2020end} uses a linear layer to predict the object classes and a three-layer MLP to regress the bounding boxes, based on $h$.
Motivated by DETR,
we use a linear layer for the prediction of movable, rigidity, articulation class and action.
We use a three-layer MLP to predict the bounding boxes and rotation axes, as they require localization.
We add a gaussian bump~\cite{law2018cornernet} for affordance ground truth, where the radius is 5.

\mypar{Balance of loss functions.}
Since we use multiple loss functions for each prediction and each loss has a different range, they need to be balanced.
We treat the weights of losses as hyperparameters and tune them accordingly.
The weights of movable, rigidity, articulation class, and action losses are 0.5. 
The weights of mask losses (both focal loss~\cite{lin2017focal} and DICE~\cite{milletari2016vnet}) are 2.0. 
The weights of box L1 loss is 5.0 and generalized IoU loss is 2.0.
The weights of axis angle loss is 1.0 and axis offset loss is 10.0.
The weights of affordance loss is 100.0.
The weights of depth losses are 1.0.
For both focal losses of segmentation masks and affordance map, we use $\gamma=2$. For the focal loss of segmentation mask, we use $\alpha=0.25$ to balance positive and negative examples. In affordance we use the standard $\alpha = 0.95$ since there are much more negatives than positives. 

\mypar{Training details.}
The image encoder, prompt encoder and the mask decoder are pretrained on Segment-Anything~\cite{kirillov2023segany}.
To save gpu memory, we use SAM-ViT-b as the image encoder, which is the lightest pretrained model.
The other heads (e.g. affordance) are trained from scratch.
We use an AdamW optimizer~\cite{loshchilov2017decoupled} of the learning rate $10^{-4}$ and train the model for 200 epochs.
The input and output resolution is 768$\times$1024.
The batch size is 2.
We train the model on four NVIDIA A40 gpu, with distributed data parallel.

\mypar{Rendering 3D Interaction.}
Given all these predictions, we are able to predict the potential 3D object interaction of articulated objects from a single image.
For articulated objects with a rotation axis, we first backproject the predicted 2D axis to 3D, based on the predicted depth~\cite{Qian22}.
We then rotate the object point cloud along the 3D axis and project it back to 2D.
We fit a homography between the rotated object points and the original one, using RANSAC~\cite{fischler1981random}.
Finally, we warp the homography on the original object mask.
There is a similar procedure for articulated objects with a translation axis.
Instead, we estimate an average surface normal of the object, and use it as the direction of translation axis~\cite{liu2018planenet,liu2019planercnn,Qian22}.
Moreover, the interaction of deformable objects is high dependent of its material, which is difficult to predict from pure visual cues~\cite{yang2022touch}.
On the other hand, freeform objects can be moved without any constraints. 
Therefore, in this paper, we only render 3D interaction for articulated objects.
We use pytorch3D~\cite{ravi2020pytorch3d} and opencv to implement the projection and homography fitting.
Final results are shown in the animation video.

\begin{figure}[t]
    \centering
    \includegraphics[width=\linewidth]{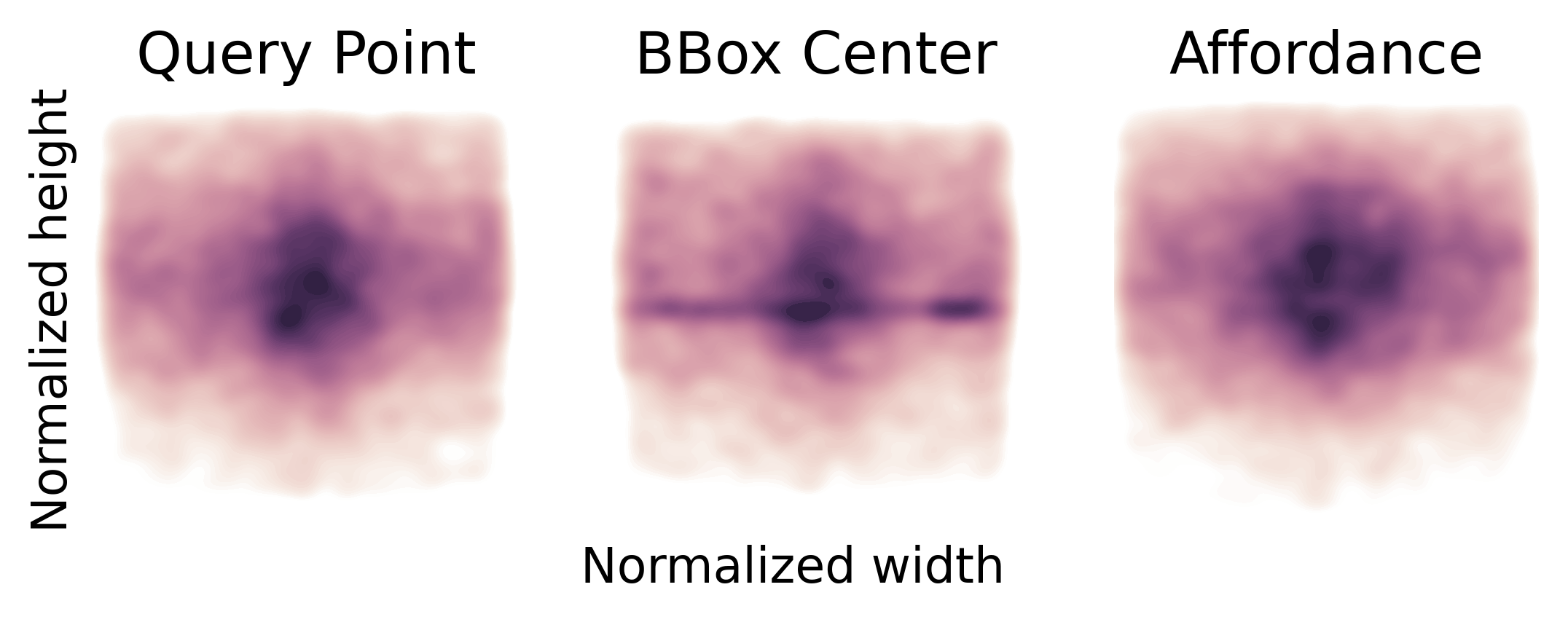}
    \includegraphics[width=\linewidth]{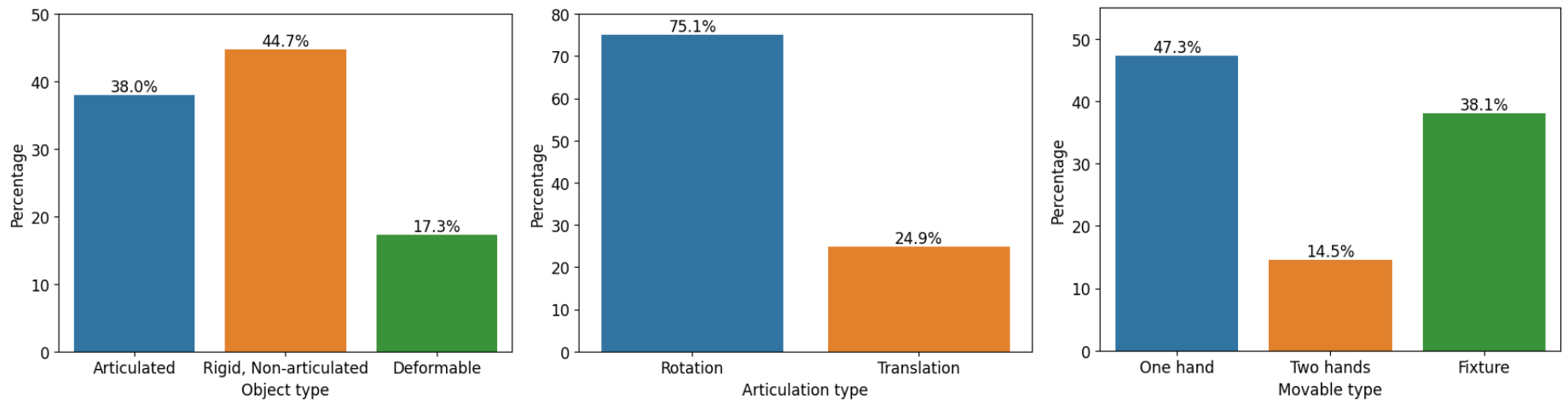}
    \caption{Statistics of our 3DOI dataset. \textbf{(Row 1)} We show the distribution of query points, box centers, and affordance in normalized image coordinates, similar to LVIS~\cite{gupta2019lvis} and Omni3D~\cite{brazil2023omni3d}. \textbf{(Row 2)} We show the distribution of object types, articulation types and movable types.}
    \label{fig:stats}
    \vspace{-1em}
\end{figure}

\begin{figure*}[t]
    \centering
     \includegraphics[width=\linewidth]{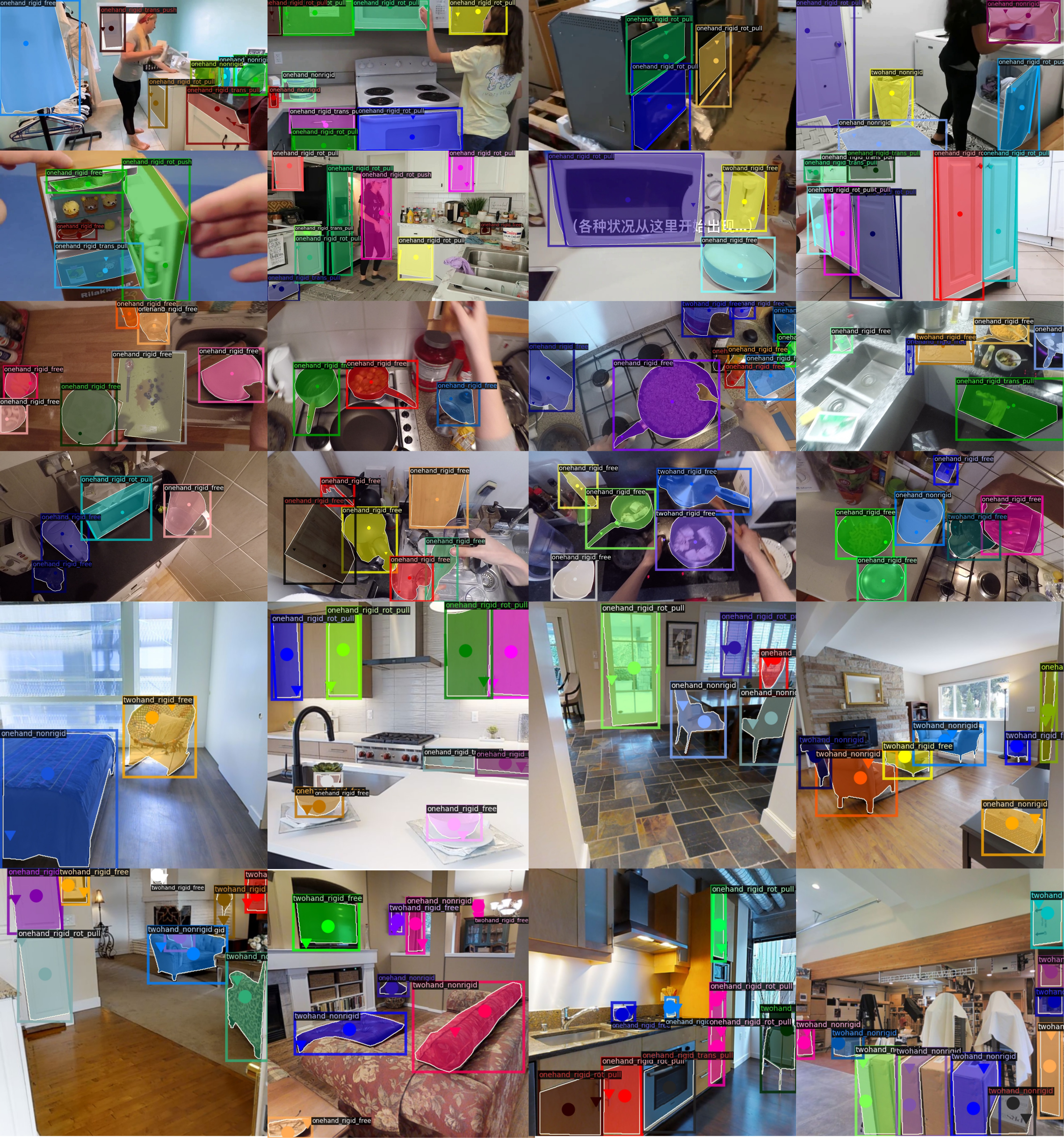}
    \caption{Example annotations of our 3DOI dataset. Row 1-2 come from Internet videos~\cite{Qian22}. Row 3-4 come from egocentric videos~\cite{damen2022Rescaling}. Row 5-6 come from renderings of 3D dataset~\cite{eftekhar2021omnidata}. \tikzcircle[black, fill=black]{2pt} is the query point, and $\blacktriangledown$ is the affordance.}
    \vspace{-1em}
    \label{fig:supp_data}
\end{figure*}

\section{Data Collection}
\label{app:data}

In this section, we introduce steps of the data annotation.
We show the statistics of our dataset in Figure~\ref{fig:stats}. 
We also show additional annotations in Figure~\ref{fig:supp_data}.

\mypar{Selecting query points.}
We first ask workers to select approximately five query points for each image.
The query point should be on an interactive object.
Some query point should be on large objects, while others should be on small objects.
We annotate more query points of fixtures later, as fixtures do not need additional annotations.

\mypar{Bounding boxes.}
According to the query point, we ask workers to draw a bounding box.
The bounding box should only cover the movable part of an object.
For example, if the query point is on the door of a refrigerator, the bounding box should only cover the door, instead of the whole refrigerator.
It is because we are asking ``what can I do here''.

\mypar{Properties of the object.}
We then annotate properties of the object.
It is a series of multiple choice questions:
(1) can the object be moved by one hand, or two hands?
(2) is the object rigid or not?
(3) if it is rigid, is it articulated or freeform?
(4) if it is articulated, is the motion rotation or translation?
(5) if we want to interact with the articulated object, should I push or pull?

\mypar{Rotation Axes.}
For objects which can be rotated, we ask workers to draw a 2D line to represent the rotation axis.

\mypar{Segmentation Masks.}
For all objects, we further ask workers to draw the segmentation mask of the movable part.

\mypar{Fixtures.}
Finally, we collect another 10K images and randomly sample 5 query points for each image.
We ask workers to annotate whether they are fixtures or not.
We mix the dataset with these annotations.

\end{document}


\title{Supplementary Material for \\Understanding 3D Object Interaction from a Single Image}

\author{
Shengyi Qian \kern15pt David F. Fouhey\\
   University of Michigan\\
	{\tt\small \{syqian,fouhey\}@umich.edu}\\
   {\small \url{https://jasonqsy.github.io/Articulation3D}}
}

\maketitle
\ificcvfinal\thispagestyle{empty}\fi

\appendix

\input{sections/supp}

{\small
\bibliographystyle{ieee_fullname}
\bibliography{local}
}